\documentclass{article}

\usepackage{PRIMEarxiv}

\usepackage[utf8]{inputenc} %
\usepackage[T1]{fontenc}    %
\usepackage{hyperref}       %
\usepackage{url}            %
\usepackage{booktabs}       %
\usepackage{amsfonts}       %
\usepackage{nicefrac}       %
\usepackage{microtype}      %
\usepackage{lipsum}
\usepackage{fancyhdr}       %
\usepackage{graphicx}  
\usepackage{hyperref}
\usepackage{xcolor}%
\usepackage{graphicx}
\usepackage{cleveref}
\usepackage{subcaption}
\usepackage[table]{xcolor}
\usepackage{multirow}
\usepackage{bbding}
\usepackage{pifont}
\usepackage{wasysym}
\usepackage{amssymb}
\usepackage[edges]{forest}
\newcommand{\xmark}{\ding{55}}%
\definecolor{myorange}{RGB}{255,127,0} %
\definecolor{Gray}{gray}{0.85} %
\newcommand{\tocite}[1]{{\color{orange} [cite something]}}
\newcommand{\todo}[1]{{\color{green} [TODO]}}
\newcommand{\checkyes}[1]{\textcolor{green}{\Checkmark}}
\newcommand{\checkno}[1]{\textcolor{red}{\xmark}}
\graphicspath{{media/}}     %

\title{African Languages AI Safety Readiness Report

}

\author{\normalsize Israel Abebe Azime$^{1,2}\thanks{Equal contribution.}$, Jesujoba O. Alabi$^{1,2*}$, Tajuddeen Gwadabe$^{1}$, David Ifeoluwa Adelani$^{1,3}$\\
\footnotesize
 $^1$Masakhane, $^2$Saarland University, \\
  \footnotesize
 $^3$Mila - Quebec AI Institute, McGill University, and Canada CIFAR AI Chair
}

\begin{document}
\maketitle

\begin{abstract}

With the widespread adoption of large language models (LLMs) across languages and domains, AI safety has emerged as a critical global concern. However, due to resource constraints, research and development efforts tend to prioritize a limited number of high-resource languages, leaving many others, especially those spoken in Africa, underrepresented. While existing research on language “resource-ness” has largely examined whether sufficient resources are available to develop language technologies, the question of whether languages are adequately represented in AI safety research has received attention only more recently. %
This report examines the readiness of African languages for inclusion in AI safety discussions, drawing on existing literature and data. It considers four factors: the availability of AI-relevant resources, including text data; national AI policy support; performance on established text benchmarks; and speaker population. 
Based on these factors, languages are grouped into four readiness categories: \textbf{fully ready}, \textbf{mostly ready}, \textbf{emerging}, and \textbf{foundational support needed}. This classification is intended to inform the inclusion of African languages in AI safety benchmarks, highlight varying levels of preparedness, and guide strategic investment in data development, policy, and community engagement for underrepresented languages.

\end{abstract}

 \keywords{AI Safety \and LLM \and Multilingual Safety}

\section{Introduction}

The emergence of large language models (LLMs) has advanced natural language processing (NLP), enabling models to understand and generate human-like text across a range of tasks and domains~\cite{qin2024large,naveed2023comprehensive}. %

Although there are over 7,000 languages spoken worldwide~\cite{campbell2008ethnologue}, mainstream NLP research has historically focused primarily on English. 
In recent years, more studies have begun to address low-resource and underrepresented languages~\cite{hedderich2020survey}. This work includes developing language-specific models~\cite{wu2020all,conneau2019unsupervised}, evaluating their performance using appropriate benchmarks~\cite{mohtashami2023learning,falcao-etal-2024-comet}, and evaluating fairness and bias in NLP systems~\cite{gallegos2024bias}.
Recent surveys of multilingual LLMs also highlight ongoing challenges in evaluating models across languages and addressing differences in their performance~\cite{Qin2025ASO}.

These differences also raise concerns about AI safety. LLMs can generate biased, misleading, or harmful content, particularly when safety safeguards are inadequate or training data do not adequately represent the languages and communities they serve~\cite{Bender2021OnTD,Weidinger2022TaxonomyOR}. These risks raise particular concerns for low-resource languages, where limited digital corpora and annotated datasets can constrain model development and adaptation~\cite{penedo2024fineweb-2}. This scarcity not only hampers the performance of AI systems in these languages but also exacerbates safety concerns~\cite{wang2023all}, as most AI safety research has primarily focused on English~\cite{Yong2025TheSO}.

In this report, we focus on answering a central question: \textit{which languages native to Africa are prime candidates for inclusion in future AI safety benchmarks?} %
To address this question, we conduct a comprehensive investigation comprising four key components. We did the following: 
\textbf{(1)} analyzed the language coverage in widely-used datasets to understand existing representation disparities; 
\textbf{(2)} evaluated popular multilingual benchmarks, examining both language inclusion and performance scores across languages;
\textbf{(3)} benchmarked state-of-the-art large language models (LLMs) using a carefully curated test suite designed to assess their capabilities in understanding and generating text in these languages; and
\textbf{(4)} conducted interviews with native speakers to gain qualitative insights into the current state of AI readiness, as well as to identify challenges, opportunities, and directions for future development and support.

\begin{figure}[]
  \centering
  \includegraphics[width=\textwidth]{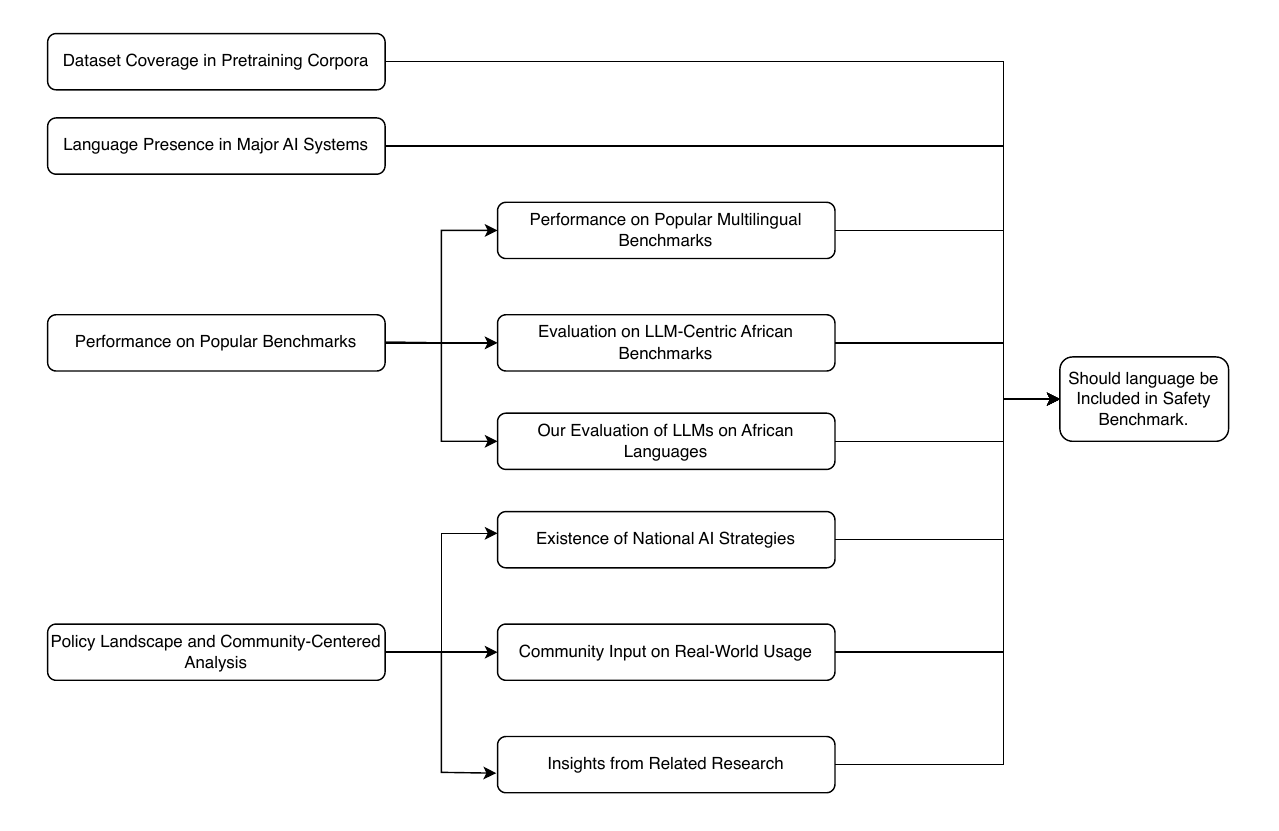} %
  \caption{A layered framework for evaluating the presence and readiness of African languages in machine learning ecosystems. The analysis spans dataset coverage in pretraining corpora, representation in major AI systems, and performance across both popular multilingual benchmarks and African-specific LLM benchmarks. It also incorporates qualitative dimensions such as national AI strategies, policy landscapes, and community-centered insights, including real-world usage and feedback from related research. This comprehensive view is used to assess not only technological readiness but also to guide the inclusion of languages in safety-critical benchmarks and broader AI resource prioritization. The result is highly dependent on Language ID used.}
  \label{fig:example}
\end{figure}

\Cref{fig:example} shows a summary of our study, showing the layered framework we used to arrive at the categorization of African languages into four levels of AI safety readiness.

\section{Background}

\subsection{Languages}
Africa has one of the richest linguistic landscapes in the world, with over 2,000 languages spoken across the continent~\cite{ethnologue2025}. These languages belong to several language families, such as Afroasiatic, Nilo-Saharan, and Niger-Congo, with the Niger-Congo family being the largest, both in terms of the number of languages and speakers~\cite{hammarstrom2024bank,ethnologue2025}. Despite this vast diversity, African languages are not well represented in language technologies. However, there are efforts from various quarters, including research institutions, and local language communities, in the field of NLP for African languages, including the development of datasets, embeddings, language models, and other essential resources~\cite{Alabi2025ChartingTL}. Nevertheless, the readiness of many African languages for integration into AI technologies remains an open research question, due to challenges such as limited data and resource availability. Therefore, in this paper we study and analyse the readiness of readiness of many African languages for safety use in AI. 

\subsection{Datasets}
Several datasets exist for African languages to support various NLP downstream tasks such as machine translation~\cite{adelani-etal-2022-thousand,Alabi2025AFRIDOCMTDM}, sentiment analysis~\cite{muhammad-etal-2023-afrisenti,muhammad-etal-2022-naijasenti}, named entity recognition~\cite{adelani-etal-2021-masakhaner,adelani-etal-2022-masakhaner}, and question answering~\cite{ogundepo-etal-2023-cross}. While some datasets are developed for individual languages~\cite{adelani-etal-2021-effect,Wanjawa2022KenSwQuADAQA,taffa-etal-2024-low}, others are part of multilingual efforts~\cite{adelani-etal-2022-thousand,muhammad-etal-2023-afrisenti,adelani-etal-2022-masakhaner}. These multilingual datasets may be created through translation of existing resources in high-resource languages~\cite{adelani-etal-2025-irokobench}, annotation projection~\cite{adelani-etal-2024-sib}, or extracting massive unannotated data from the web~\cite{xue-etal-2021-mt5,penedo2024fineweb-2,kudugunta2023madlad400}, which are often used for language modeling.~\cite{Alabi2025ChartingTL}, in their work, showed that efforts on NLP tasks for African languages are skewed toward popular tasks, which can be attributed to the availability of datasets.

\subsection{Multilingual Language Models}
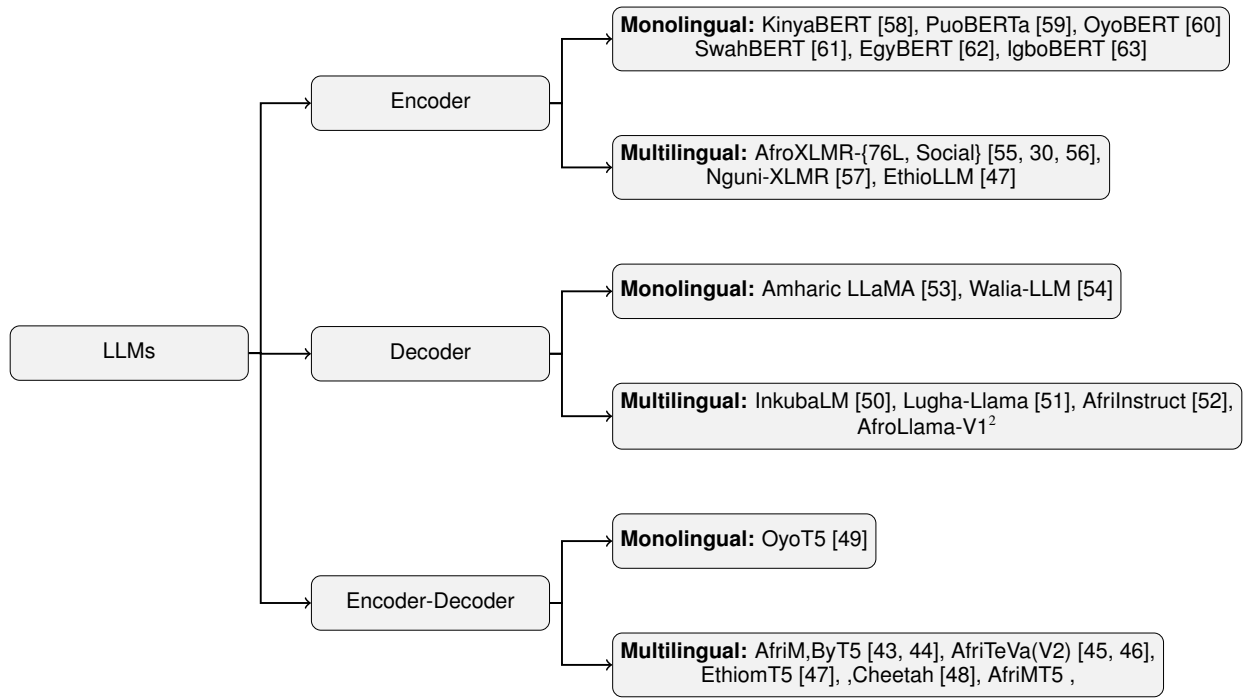
\begin{figure}[htbp]
\centering
\scalebox{0.9}{ %
\begin{forest}
for tree={
    grow=east,
    draw,
    rounded corners,
    parent anchor=east,
    child anchor=west,
    edge={->, thick},
    anchor=west,
    edge path={
        \noexpand\path [draw, \forestoption{edge}]
        (!u.parent anchor) -- +(5pt,0) |- (.child anchor)\forestoption{edge label};
    },
    font=\sffamily\small,
    l sep+=15pt,
    s sep+=20pt,
    fill=gray!10,
    minimum height=0.8cm,
    align=center,
    if level=0{
        minimum width=3.5cm,  %
    }{
        if level=3{
            text width=3cm,    %
            align=left,        %
        }{
            minimum width=3.5cm, %
        },
    },
}
[LLMs
  [Encoder-Decoder
    [{\textbf{Multilingual:} AfriM,ByT5~\cite{ogueji-etal-2021-small,oladipo2024scaling}, AfriTeVa(V2)~\cite{jude-ogundepo-etal-2022-afriteva,oladipo-etal-2023-better}, \\ EthiomT5~\cite{tonja2024ethiollm}, ,Cheetah~\cite{adebara-etal-2024-cheetah}, AfriMT5 , }] 
    [{\textbf{Monolingual:} OyoT5~\cite{olawole2024yadleveragingt5improved}}] 
  ]
  [Decoder 
    [{\textbf{Multilingual:} InkubaLM~\cite{tonja2024inkubalmsmalllanguagemodel}, Lugha-Llama~\cite{Buzaaba2025LughaLlamaAL}, AfriInstruct~\cite{uemura-etal-2024-afriinstruct}, \\
    AfroLlama-V1\footnote{\url{https://huggingface.co/Jacaranda/AfroLlama_V1}}
    }]
    [{\textbf{Monolingual:} Amharic LLaMA~\cite{andersland2024amharicllamallavamultimodal}, Walia-LLM~\cite{azime-etal-2024-walia}}] 
    ]
  [Encoder
    [{\textbf{Multilingual:} AfroXLMR-\{76L, Social\}~\cite{alabi-etal-2022-adapting,adelani-etal-2024-sib,belay2025afroxlmrsocialadaptingpretrainedlanguage}, \\
    Nguni-XLMR~\cite{meyer-etal-2024-nglueni}, EthioLLM~\cite{tonja2024ethiollm}}]  %
    [{\textbf{Monolingual:} KinyaBERT~\cite{nzeyimana-niyongabo-rubungo-2022-kinyabert}, PuoBERTa~\cite{marivate2023puoberta}, OyoBERT~\cite{david_adelani_2023} \\ SwahBERT~\cite{martin-etal-2022-swahbert},  EgyBERT~\cite{qarah2024egybert}, IgboBERT~\cite{chukwuneke-etal-2022-igbobert}}]
  ]
]
\end{forest}
}
\caption{Classification of African-centric Large Language Models (LLMs) based on their base transformer architecture.}
\label{fig:llm_tree}
\end{figure}

With the curation of massive multilingual resources from the web, recent language models now cover hundreds of languages, including many African languages~\cite{conneau-etal-2020-unsupervised,lin-etal-2022-shot,kudugunta2023madlad400}. These models follow one of three transformer architectures~\cite{Vaswani2017AttentionIA}, encoder-only~\cite{conneau-etal-2020-unsupervised}, decoder-only~\cite{lin-etal-2022-shot,kudugunta2023madlad400}, and encoder-decoder~\cite{xue-etal-2021-mt5}. \Cref{fig:llm_tree} shows the categorization of African-centric LLMs based on these architecture. While some of these models are trained from scratch, others are developed through continued pretraining.  Recently, decoder-only models have become more common, contributing to a shift in how NLP tasks are formulated~\cite{Dubey2024TheL3,Brown2020LanguageMA,raffel2020exploring}. Tasks that were once treated as classification or structured prediction problems are now increasingly rendered as generative tasks. As these models continue to scale in size, they have demonstrated impressive reasoning capabilities  and emergent abilities~\cite{Brown2020LanguageMA,wei2022emergent}, skills that were not explicitly programmed but arise from model complexity and size. However, this rapid advancement also raises significant concerns about their safety and reliability.. 

Furthermore, many of the recent and popular LLMs are trained predominantly on English data~\cite{Achiam2023GPT4TR,team2025gemma,team2024gemini}, with other languages, particularly low-resource ones, being underrepresented. Nevertheless, some have shown an ability to generalize across languages. Also, there are growing efforts to develop models specifically for African languages, either by training from scratch using curated local datasets or by adapting existing multilingual models through continued pretraining.

\section{Methodology}

To assess the AI readiness of African languages, we developed a multi-faceted methodology that draws on existing data, quantitative analysis, and community insights. In this work, we consider languages to be 'ready' when they have accessible corpora, pretrained models, established benchmarks, supportive national AI policies, and an active community of speakers, used in this report as a proxy for societal importance. Each subsection below describes one of the core components of our approach.

\subsection{Dataset Coverage in Pretraining Corpora}
The development of language technologies based on deep learning and transformer models requires large volumes of data. In particular, modern language models, which underpin a wide range of NLP applications, depend heavily on large text corpora, typically sourced from the web. To assess the AI safety readiness of African languages, we began by examining their representation in large-scale pretraining datasets, focusing on recent massively multilingual corpora such as MADLAD-400 \cite{kudugunta2023madlad400} and FineWeb-2 \cite{penedo2024fineweb-2}, and Wikipedia. These corpora are among the most widely used in the training of contemporary multilingual language models. For each language, we extracted the number of tokens, sentences, documents, or articles, depending on what the dataset provided. We then compared these amounts to English and other widely used languages. This allowed us to gauge the extent of exposure each language receives during pretraining, a key factor influencing downstream model performance.

\subsection{Language Presence in Major AI Systems}

To assess global inclusion efforts, we analyzed the presence of African languages in widely-used AI systems such as Google Translate\footnote{https://translate.google.com/} and Meta’s NLLB~\cite{costa2022no}. These systems serve as proxies for practical deployment readiness and the extent to which a language is accessible via commercial or research-grade models. We documented whether the language is supported, and the performance it has.

\subsubsection{Model Support and Benchmarking}
To understand how African languages perform in standard multilingual evaluation settings, we analyzed results reported in~\cite{Beyene2025mSTEBMM}, which evaluated LLMs on benchmarks such as FLORES-200~\cite{nllb2022} for machine translation, SIB-200~\cite{adelani-etal-2024-sib} for topic classification, and BELEBELE~\cite{bandarkar-etal-2024-belebele} for reading comprehension based question answering, covering over 100 languages. We examined metrics such as chrF++~\cite{popovic-2015-chrf}, BLEU~\cite{papineni-etal-2002-bleu}, and accuracy. We also compared the performance of African languages to languages from other world regions to contextualize their relative standing. In addition, we used results from~\cite{ojo2025afrobenchgoodlargelanguage}, which introduced AfroBench, a collection of benchmark tasks specifically focused on African languages to further analyze model performance across tasks such as sentiment analysis, named entity recognition, and question answering. These comparisons allowed us to identify languages that are underperforming despite moderate representation and those that show strong results even with limited resources.

Languages that achieve stronger performance on these benchmarks can be considered more ready for AI applications, as they are better supported by current models. Benchmark performance serves as a proxy for model usability in real-world tasks and reflects how well the language has been represented and learned during pretraining.

\subsection{Policy Landscape and Community-Centered Analysis}

\subsubsection{Existence of National AI Strategies}

In this report, we view the presence of a national AI policy as an indicator of potential for regulatory oversight, investment in local capacity, and the development of infrastructure needed to monitor and guide the deployment of AI technologies. Consequently, languages spoken in countries with such policies may be better positioned for safe and responsible AI integration. Hence, using online sources, we identified African languages spoken in at least one country with a national AI policy.

\subsubsection{Community Input on Real-World Usage}

We complemented our technical analysis with qualitative insights gathered through forms filled by native speakers and local technologists. These community responses helped contextualize the technical findings, offering views on language use in digital spaces, AI-related challenges, and perceived progress in AI inclusion. Responses were thematically coded to identify recurring patterns and challenges.

\subsubsection{Insights from Related Research}

A considerable body of research has been dedicated to assessing language resources and capabilities, with the aim of categorizing languages based on their level of support and identifying the efforts required to ensure equitable technological inclusion. Notable examples of such work include~\cite{joshi-etal-2020-state,nigatu2024zeno}. This work defines what it means for a language to be low-resource by carefully analyzing sociopolitical, resource-related, and artifact-based factors. Even though we share good subset of factors to look at languages, the aim of this paper is compiling several factors to decide on which subset of languages to build our safety benchmark.

\subsection{Criteria for Language Inclusion}

As an outcome of this work, we used the inputs from the sections above to classify languages into four readiness levels: \textbf{fully ready}, \textbf{mostly ready}, \textbf{emerging}, and \textbf{foundational support needed} for integration into AI safety research and benchmarks. Based on our analysis, languages not included in FLORES-200 are categorized as \textbf{foundational support needed}, with the exception of Nigerian Pidgin, which, according to~\cite{ethnologue2025}, is the most widely spoken language in Africa and a creole known for high cross-lingual transfer. As a result, the 57 African languages included in FLORES-200, along with Nigerian Pidgin, were ranked into the remaining three readiness categories.

\begin{figure}[htbp]
  \centering
  \begin{subfigure}[t]{0.45\textwidth}
    \includegraphics[width=\linewidth]{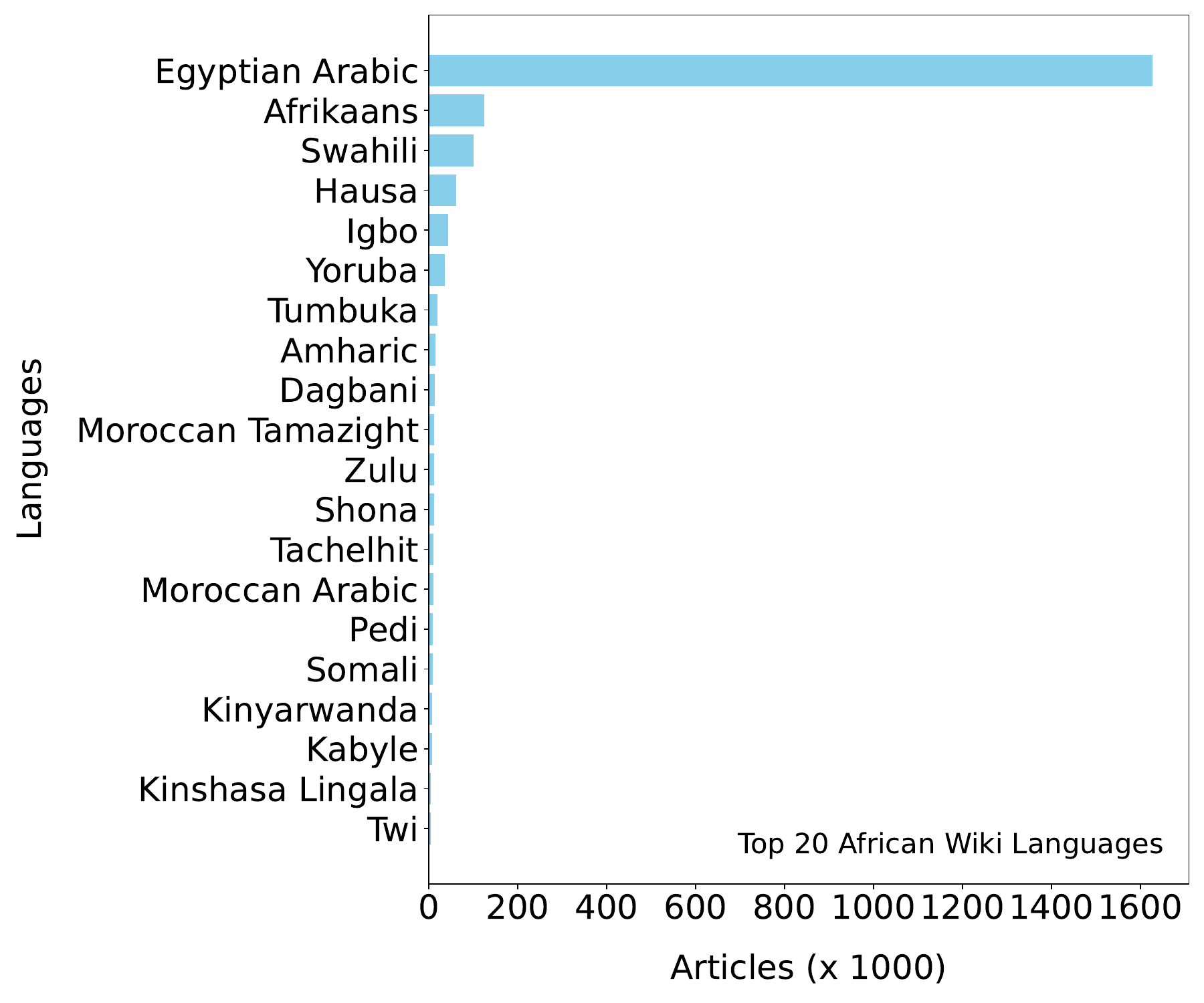}
    \caption{Wikipedia}
    \label{fig:image1}
  \end{subfigure}
  \hfill
  \begin{subfigure}[t]{0.45\textwidth}
    \includegraphics[width=\linewidth]{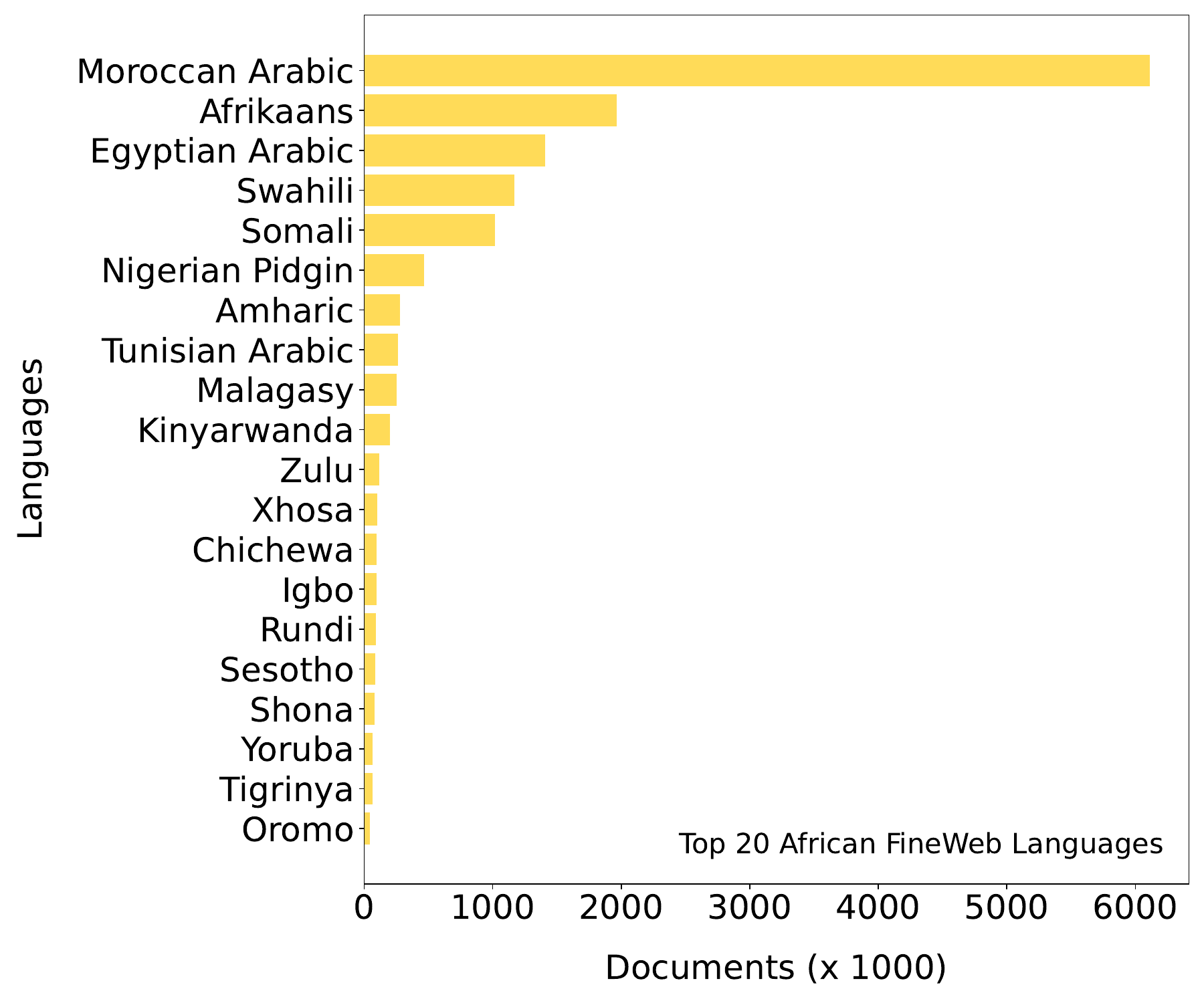}
    \caption{FineWeb}
    \label{fig:image2}
  \end{subfigure}
  \caption{Top 20 African languages in publicly available multilingual unlabeled corpora: Wikipedia and FineWeb.}
  \label{fig:side_by_side}
\end{figure}

\section{Result and Analysis}
Based on the described methodology, we proceed to answer a series of sub-questions that collectively address the overall research question: which African languages are ready for AI safety?

\subsection{What languages are represented in existing Corpora?}
\label{sec:corpora}

Our analysis of Wikipedia statistics\footnote{retrived on 06.06.2025 from  \href{https://en.wikipedia.org/wiki/List_of_Wikipedias}{List of Wikipedias}.} shows that of 341 active languages on Wikipedia, only 53 are African languages. Furthermore, of the more than 65 million articles, just over 10\% are in English, German accounts for 4.6\%, while all 53 African languages combined account for less than 5\%.

Aside from Wikipedia, there have been efforts to curate massive unannotated corpora for several languages, including African languages, from the entire web. Recent datasets include mC4~\cite{xue-etal-2021-mt5}, Fineweb2~\cite{penedo2024fineweb-2}, and MADLAD~\cite{kudugunta2023madlad400}. Our analysis of the Fineweb2 dataset statistics provided by its authors, excluding English, shows that out of over 4 billion documents, 605 million (13.3\%) are in Russian, while 332 million (7\%) are in French. In contrast, African languages make up less than 0.5\% of the dataset.

\begin{table*}[th]
\centering
\setlength\tabcolsep{4pt}
\resizebox{\linewidth}{!}{%
\begin{tabular}{lrrr|rrr|rrr|rrr|rrr}
\toprule
 & \multicolumn{3}{c|}{\textsc{LID (acc $\uparrow$)}} &  \multicolumn{3}{c|}{\textsc{Topic class. (acc $\uparrow$)}}  &  \multicolumn{3}{c|}{\textsc{RC-QA (acc $\uparrow$)}}  &   \multicolumn{3}{c|}{\textsc{Machine Transl. (ChrF++ $\uparrow$)}} & \multicolumn{3}{c}{\textsc{Ave. (class. tasks)}}\\
\textbf{Region} & \rotatebox{60}{\textbf{Gemma}} &  \rotatebox{60}{\textbf{GPT-4o}} & \rotatebox{60}{\textbf{Gemini}} & \rotatebox{60}{\textbf{Gemma}} &  \rotatebox{60}{\textbf{GPT-4o}} & \rotatebox{60}{\textbf{Gemini}} & \rotatebox{60}{\textbf{Gemma}} &  \rotatebox{60}{\textbf{GPT-4o}} & \rotatebox{60}{\textbf{Gemini}} & \rotatebox{60}{\textbf{Gemma}} &  \rotatebox{60}{\textbf{GPT-4o}} & \rotatebox{60}{\textbf{Gemini}} & \rotatebox{60}{\textbf{Gemma}} &  \rotatebox{60}{\textbf{GPT-4o}} & \rotatebox{60}{\textbf{Gemini}} \\
\midrule				
Africa &62.9 &69.6 &80.0 &59.9 &69.6 &74.9 &60.6 &70.2 &77.0 &34.8/24.0 &41.0/30.5 &46.2/32.0 &61.4 &69.6 &77.5 \\
Americas/Oceania &79.2 &99.6 &99.5 &69.5 &80.5 &82.9 &62.3 &78.2 &80.7 &41.2/32.4 &49.0/40.8 &52.6/42.9 &74.3 &90.0 &91.2 \\
Asia (S) &78.2 &83.2 &87.2 &81.5 &82.1 &86.7 &78.6 &83.0 &84.7 &51.9/33.3 &54.2/34.2 &59.6/39.2 &79.8 &82.7 &87.0 \\
Asia (SE) &76.2 &82.0 &86.6 &75.8 &80.9 &84.5 &75.5 &76.3 &81.7 &48.4/35.1 &51.8/38.5 &57.6/42.1 &76.0 &81.5 &85.5 \\
Asia (W, C) &99.3 &99.2 &99.5 &85.3 &86.8 &87.7 &83.3 &87.0 &87.5 &56.5/38.8 &58.6/44.3 &60.8/45.9 &92.3 &93.0 &93.6 \\
Asia (E) &99.7 &100.0 &99.8 &85.0 &87.2 &86.7 &87.7 &90.2 &85.9 &54.5/29.7 &56.3/33.7 &57.4/32.8 &92.3 &93.6 &93.3 \\
Europe (W, N, S) &94.2 &98.4 &97.9 &85.1 &87.5 &88.2 &90.3 &92.6 &87.2 &62.9/52.7 &64.9/56.5 &66.2/56.9 &89.7 &93.0 &93.1 \\
Europe (E) &90.0 &95.7 &97.8 &85.0 &87.5 &88.0 &89.2 &92.8 &86.2 &60.3/48.5 &62.4/49.2 &63.9/51.1 &87.5 &91.6 &92.9 \\

\midrule       
Average (macro)     & 85.0 & 91.0           & \textbf{93.5}  & 78.4 & 82.8          & \textbf{85.0}   & 78.4 & 83.8   & \textbf{83.9} & 51.3/36.8 & 54.8/41.0 & \textbf{58.0/42.9} & 78.0 & 82.5 & \textbf{83.4} \\ 
Average (micro)     & 81.0 & 86.4           &\textbf{90.5}  & 75.8 & 80.6          & \textbf{83.5} & 77.5  &82.3   & \textbf{83.4} &49.7/36.4   &53.3/40.7 &\textbf{57.0/42.7} &   \\

\bottomrule
\end{tabular}
}
\caption{Average text-based evaluation results across languages by region, using Gemma (3 27B), Gemini-2-Flash, and GPT-4o (08/24). Results obtained from \cite{Beyene2025mSTEBMM}}
\label{tab:overall_by_region_text}

\end{table*}

\Cref{fig:side_by_side} shows the top 20 African languages according to the number of Wikipedia articles and Fineweb2 documents. Even among African languages, there is a skewed distribution of available web resources. Arabic dialects (Egyptian and Moroccan) have significantly more content on Wikipedia and Fineweb2, respectively, than other languages. Afrikaans consistently ranks second across both data sources, while Swahili, the most widely spoken indigenous African language, holds the third and fourth positions, respectively.

\subsection{How good are popular LLMs on existing benchmark?}
In \Cref{sec:corpora}, we showed that web-based resources such as Wikipedia and FineWeb-2 are skewed, with African languages significantly underrepresented. In this section, we analyze and evaluate several existing LLMs with a focus on African languages. We begin with a global perspective before narrowing our evaluation to concentrate specifically on African languages.

\subsubsection{On a global perspective}
\label{sec:global_res}
We analyze the multilingual evaluation results across two broad categories of tasks: natural language understanding (NLU) and natural language generation (NLG), using the results from~\cite{Beyene2025mSTEBMM}. Specifically, we focus on language identification, topic classification, and reading comprehension-based question answering for NLU, and machine translation for NLG. All tasks are based on the massively multilingual FLORES-200 dataset~\cite{nllb2022}, which covers 200 languages. FLORES-200 was used to create SIB-200~\cite{adelani-etal-2024-sib} for topic classification, BELEBELE~\cite{bandarkar-etal-2024-belebele} for reading comprehension question answering (available for 122 languages), and also served as the benchmark for machine translation evaluation.

The results, broken down by region and presented in \Cref{tab:overall_by_region_text}, compare three popular LLMs across these tasks. On average, the findings show that African languages consistently achieve the lowest performance across all models. A closer look at the individual tasks confirms this same trend. We hypothesize that this underperformance is partly due to the underrepresentation of African languages during the pretraining and adaptation phases of these LLMs.

Additionally, proprietary LLMs tend to outperform open-source models. While the reasons for this disparity are not entirely clear, it is possible that the companies behind proprietary models have access to higher-quality or more extensive training data.

\begin{table*}[t]
\centering
\scriptsize  
\begin{tabular}{cclccccccc|c}
\toprule
    & & & & & & & & & \textbf{MT} &  \\
    \textbf{Model} & \textbf{Size} & \textbf{Lang} & \textbf{Intent} & \textbf{TC}  & \textbf{NLI} & \textbf{RC}  & \textbf{MMLU}  & \textbf{Math}  & \textit{en/fr-xx} & \textbf{AVG} \\
    \midrule

    \rowcolor{Gray}
    \cellcolor{white}\multirow{2}{*}{Lugha-Llama} 
    & \cellcolor{white}\multirow{2}{*}{8B} 
    & \textit{eng} & 16.7 & 43.6 & 46.8 & 22.4 & 31.8 & 6.4 & 51.3 & 31.3 \\
    & & \textit{africa} & 4.1 & 34.1 & 36.7 & 23.0 & 25.2 & 1.8 & 22.1 & 21.0 \\
    \midrule
     
    \rowcolor{Gray}
    \cellcolor{white}\multirow{2}{*}{Gemma 2} 
    & \cellcolor{white}\multirow{2}{*}{9B} 
    & \textit{eng} & 36.3 & 82.5 & 70.7 & \underline{93.7} & 69.8 & 68.8 & 67.9 & 70.0 \\
    & & \textit{africa} & 27.8 & 64.0 & 40.9 & 49.3 & 36.1 & 21.7 & 37.2 & 39.6 \\ 
    \midrule

    \rowcolor{Gray}
    \cellcolor{white}\multirow{2}{*}{Gemma 2} 
    & \cellcolor{white}\multirow{2}{*}{27B} 
    & \textit{eng} & 84.0 & 89.3 & 67.8 & 93.4 & 75.6 & 85.6 & 68.5 & 80.6\\ 
    & &  \textit{africa} & 31.4& 66.6 & 43.7 & 52.1 & 40.8 & 30.6 & 39.1 & 43.5\\ 
    \midrule

    \rowcolor{Gray}
    \cellcolor{white}\multirow{2}{*}{LLaMa 4} 
    & \cellcolor{white}\multirow{2}{*}{405B} 
    & \textit{eng} & \underline{88.9} & 84.8 & 49.2 & 25.0 & 11.2 & 97.6 & 73.0 & 61.4 \\ 
    & & \textit{africa} & 73.9 & 80.6 & 45.5 & 24.6 & 15.8 & 65.0 & 42.8 & 49.7 \\ 
    \midrule

    \rowcolor{Gray}
    \cellcolor{white}\multirow{2}{*}{Gemma 3} 
    & \cellcolor{white}\multirow{2}{*}{27B} 
    & \textit{eng} & 79.6 & 87.3 & 65.5 & 93.4 & 74.2 & 87.6 & 68.9 & 79.5\\
    & &  \textit{africa} & 55.2 & 74.2 & 51.2  & 62.4 & 44.4 & 47.5 & 33.1 & 52.6\\ 
    \midrule

    \rowcolor{Gray}
    \cellcolor{white}\multirow{2}{*}{Gemini 2.0} 
    & \cellcolor{white}\multirow{2}{*}{Flash} 
    & \textit{eng} & 87.6 & 86.8 & 87.0 & 63.0 & 80.8 & 92.8 & \underline{73.1} & 79.7 \\
    & &  \textit{africa} & 82.5 & \textbf{84.9} & 66.5 & 56.8 & 57.8 & \textbf{67.5} & \textbf{49.6} & 66.5\\ 
    \midrule
    
    \rowcolor{Gray}
    \cellcolor{white}\multirow{2}{*}{GPT-4.1} 
    & \cellcolor{white}\multirow{2}{*}{--} 
    & \textit{eng} & 87.8 & \underline{89.7} & 88.5 & 73.9 & 71.4 & 82.4 & \underline{73.1} &  81.0 \\
    & &  \textit{africa} & \textbf{84.4} & 84.8 & \textbf{67.5} & 64.8 & 60.2 & 59.9 & 47.3 & \textbf{67.0} \\

    \bottomrule
\end{tabular}

\caption{LLM performance on AfroBench-Lite, covering 14 African languages across 7 tasks: intent classification (Intent), topic classification (TC), natural language inference (NLI), reading comprehension (RC), knowledge base question answering (MMLU), mathematical reasoning (Math), and machine translation (MT). Results are obtained from~\cite{ojo2025afrobenchgoodlargelanguage}.}
\label{tab:afrobench-lite-new}
\end{table*}

\begin{figure}[htbp]
  \centering
  \begin{subfigure}[t]{0.49\textwidth}
    \includegraphics[width=\linewidth]{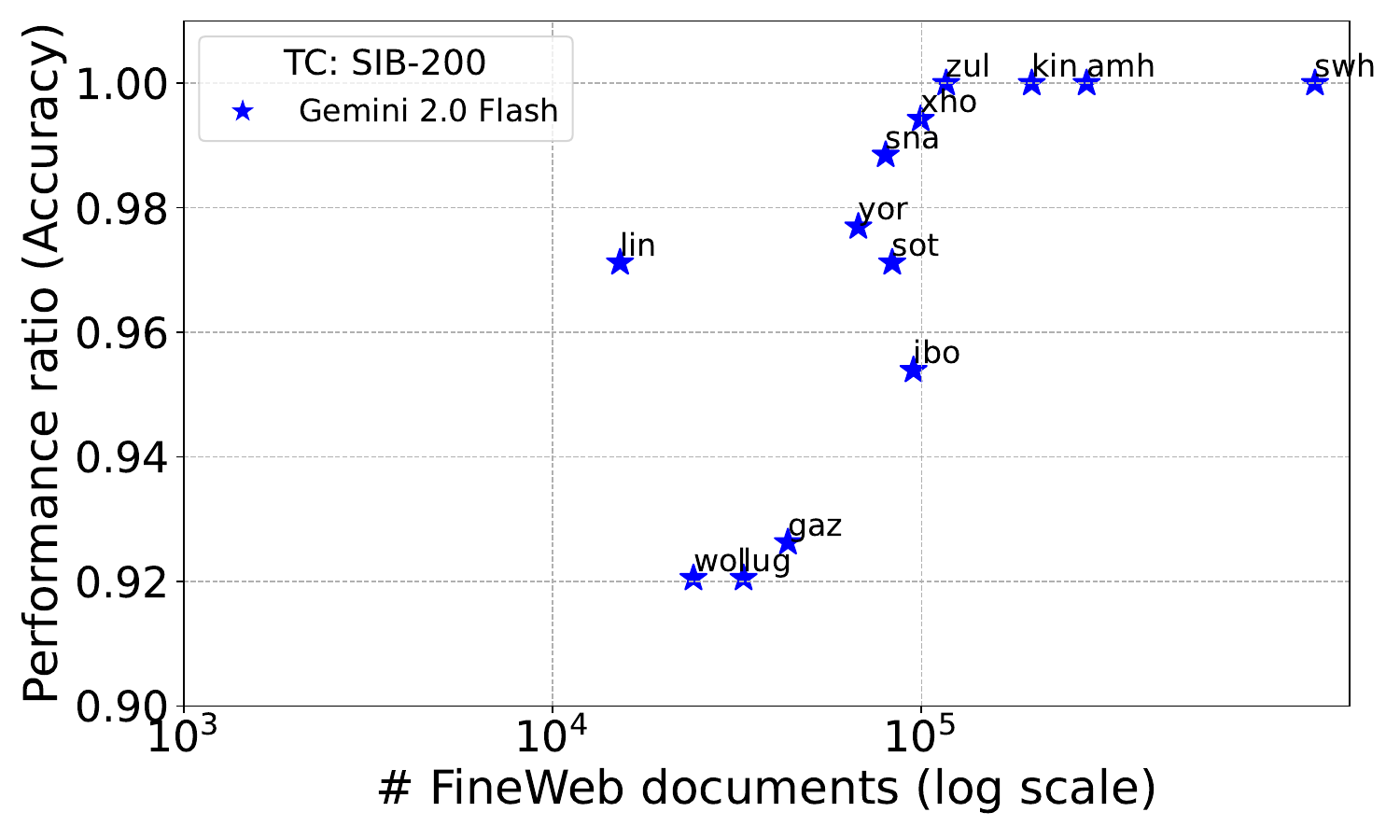}
    \caption{SIB-200}
    \label{fig:image1}
  \end{subfigure}
  \hfill
  \begin{subfigure}[t]{0.49\textwidth}
    \includegraphics[width=\linewidth]{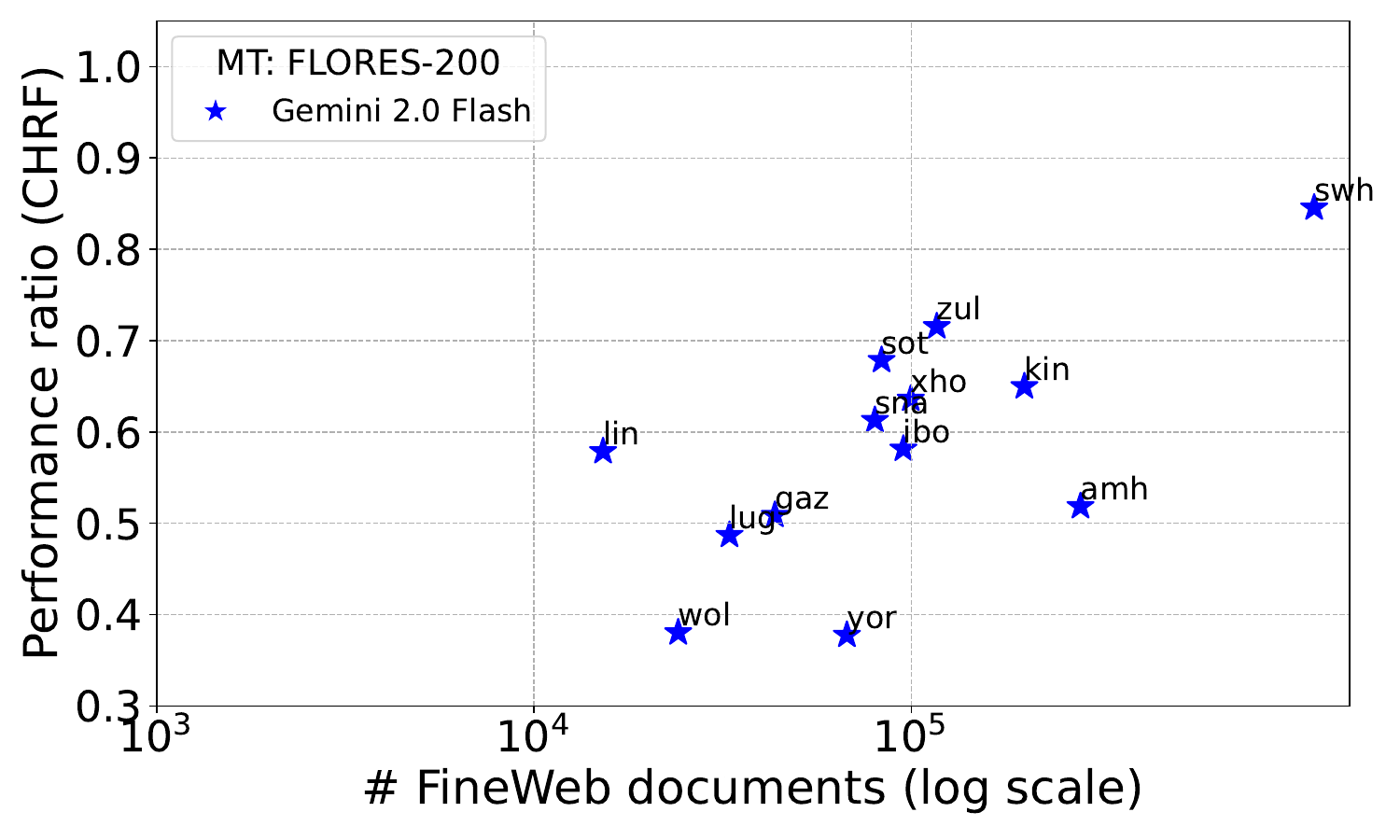}
    \caption{MT}
    \label{fig:image2}
  \end{subfigure}
  \caption{Correlation Between Data Size and Performance for African Languages.
(Left) Accuracy ratio on SIB-200; (Right) chrF++ ratio on FLORES-200. Both plots compare performance against the amount of pretraining data in Fineweb2.}
  \label{fig:correlation}
\end{figure}

\subsubsection{A focus on only African languages}

In addition to the analysis in \Cref{sec:global_res}, we evaluate the performance of several LLMs on AfroBench-Lite~\cite{ojo2025afrobenchgoodlargelanguage}, a benchmark that spans 7 tasks across 14 diverse African languages\footnote{English, Kiswahili, Kinyarwanda, Hausa, Amharic, isiXhosa, chiShona, isiZulu, Igbo, Yorùbá, Sesotho, Lingala, Oromo, Luganda, and Wolof.}. The tasks include intent classification (Intent), topic classification (TC), natural language inference (NLI), reading comprehension (RC), knowledge base question answering (MMLU), mathematical reasoning (Math), and machine translation (MT). While there is some overlap between AfroBench-Lite and the mSTEB benchmark, this analysis focuses specifically on African languages and a broader set of tasks. Accordingly, in this section, we compare the performance of 9 LLMs using results reported for 7 models by~\cite{ojo2025afrobenchgoodlargelanguage}\footnote{\url{https://mcgill-nlp.github.io/AfroBench/leaderboard.html}} along with our own experiments on 2 additional models. The results are presented in \Cref{tab:afrobench-lite-new}. 

The results show that Gemini 2.0 and GPT-4.1 perform relatively well on both English and, on average, across African languages. Across all evaluated models, performance on English is consistently higher than the average performance on African languages. The translation task, particularly translating from English or French into African languages, emerges as notably more challenging. This is expected, as machine translation is inherently more challenging, especially for low-resource African languages, because it requires both the understanding and generation capabilities of language models, unlike the other six tasks. Therefore, in our final language selection, we prioritize translation performance as a key indicator of readiness.

A closer examination of individual language performance in Gemini 2.0 reveals that Swahili, Zulu, Kinyarwanda, and Amharic are among the top-performing African languages in the topic classification task, achieving results comparable to English. In contrast, Wolof, Luganda, and Oromo exhibited the lowest performance. As previously hypothesized, this disparity is likely due to representation bias: these models are primarily English-centric, and languages with greater resource availability tend to perform better.

For the machine translation task, Swahili again outperformed other languages by a notable margin. However, Amharic showed a significant drop in performance, which we hypothesize may be due to tokenization challenges associated with its unique script. Yoruba also demonstrated relatively low performance, similar to Wolof. We attribute this to known issues with the FLORES dataset~\cite{abdulmumin-etal-2024-correcting}, which has been shown to extend to other datasets derived from it~\cite{Alabi2024AfriHuBERTAS}.

To further explore the role of representation bias, we evaluated the correlation between model performance and language resource availability. Specifically, we compared the proportion of FineWeb resources relative to English with performance on topic classification and machine translation tasks using GPT-4o, across the 14 languages in AfroBench-Lite. The results, shown in \Cref{fig:correlation}, provide partial support for the hypothesis. Swahili which is the most resourced language among the 14, consistently achieves the highest performance. Mid-resource languages such as Zulu and Hausa follow, while low-resource languages like Wolof show the weakest results. Given the limited sample size, we computed Spearman’s rank correlation between relative performance (compared to English) and FineWeb2 language size. Although the results suggest a positive trend (0.85 for SIB and 0.64 for MT), we interpret them with caution due to the small number of data points.

\begin{figure}[!]
  \centering
  \includegraphics[width=0.75\textwidth]{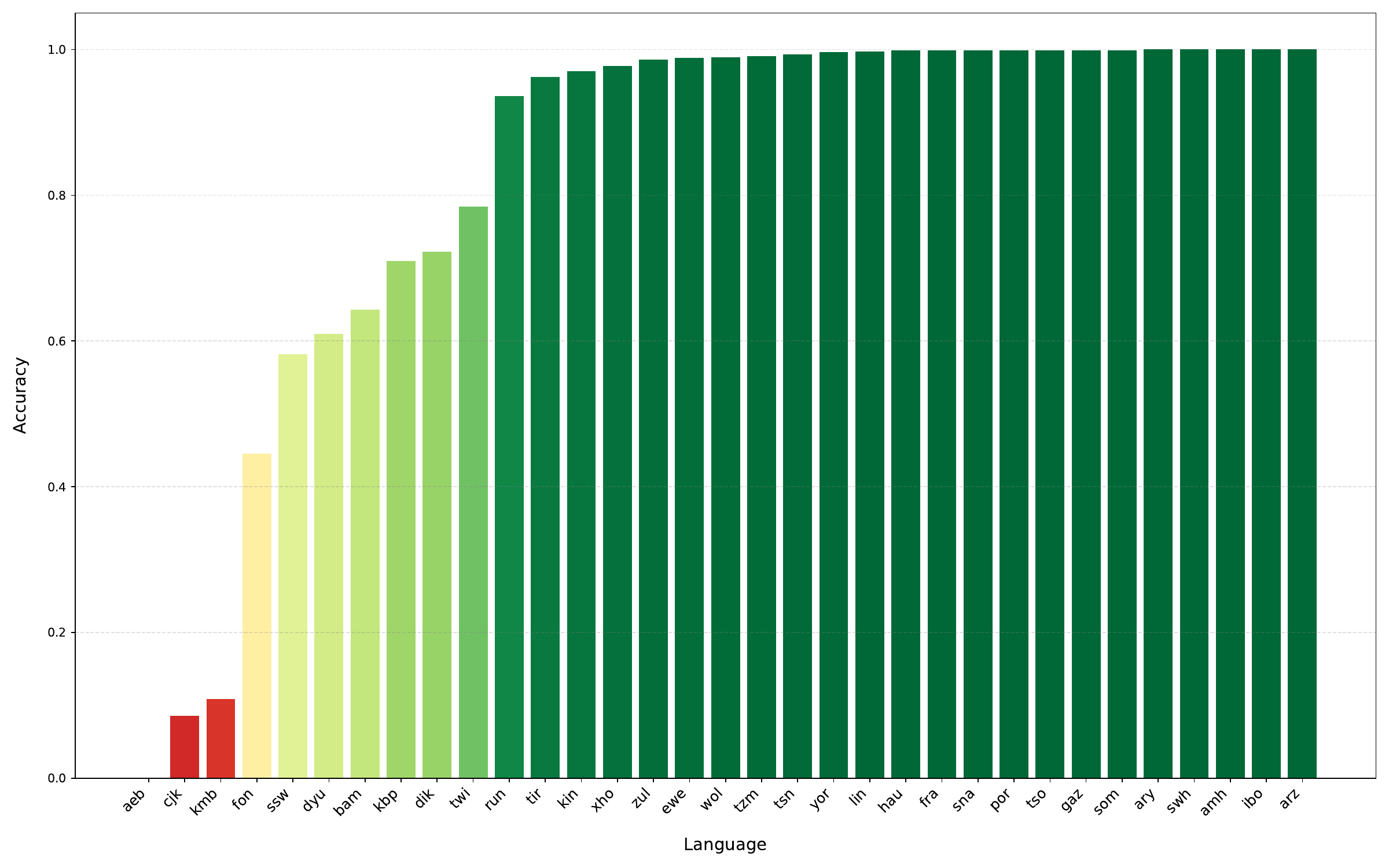} %
  \caption{Variation in language understanding accuracy of GPT-4o when prompted in different African languages from the FLORES-200 benchmark. Higher scores indicate successful comprehension and response in the correct language, while lower scores highlight failure modes such as refusal or defaulting to English. The results underscore performance disparities linked to resource availability and benchmark representation. }
  \label{fig:refusal}
\end{figure}

Figure \ref{fig:refusal}, visualizes the language understanding capability of an LLM by measuring its refusal rate on the FLORES task. To assess this, we prompted the LLM in various languages and used automatic language identification to verify if the response was in the same language as the input. A high accuracy indicates that the model correctly understood and responded in the intended language, while a low score suggests that the model either refused, responded in another language (typically English), or misunderstood the prompt. This method serves as a practical proxy for identifying whether the LLM can meaningfully engage with a given language. Our findings reveal that while the LLM accurately responds in high-resource languages like French, Portuguese, and Swahili, it frequently fails or defaults to English for many low-resource African languages such as Fon, Sotho, Sswati, Tigrinya, Oromo, and Bambara, indicating substantial gaps in the model’s multilingual capabilities, especially for languages with limited pretraining exposure or benchmark representation.

\subsection{Performance and Language Presence in Major AI Systems}
As shown in Figure \ref{fig:policy_and_nllbgc}, we present a visual comparison of the translation performance of NLLB and Google Translate across several target languages, plotted against the amount of data available from the Fineweb2 dataset. This comparison allows us to assess whether major AI systems accurately and reliably represent these languages in machine translation tasks. Initial observations from the plot suggest a general correlation: languages with larger character-level datasets tend to yield higher translation quality scores.

The under-representation of certain languages in major AI systems has significant downstream implications. In particular, it hinders equitable access to information and knowledge for speakers of these languages. With regards to safety the first approach of making systems safe is by leveraging machine translated dataset for development or testing. so the fact that the MT scores are low also makes this languages vulnerable to safety and performance issues that are not discovered while using highly resource languages.

\subsection{Which languages are spoken in countries with a national AI policy?} 
According to ~\cite{lawyershub2025privacy}, only 12 out of the 54 African countries, just 20\% have a national AI policy. These countries include Tunisia, Algeria, and Egypt in \textbf{North Africa}; Benin, Ghana, Nigeria, and Senegal in \textbf{West Africa}; Ethiopia, Mauritius, and Rwanda in \textbf{East Africa}; and South Africa and Zambia in \textbf{Southern Africa}. Meanwhile, Kenya and Morocco are currently formulating their own policies, with Kenya’s draft already available for public review.

\begin{figure}[htbp]
  \centering
  \begin{subfigure}[t]{0.4\textwidth}
    \centering
    \includegraphics[height=6cm]{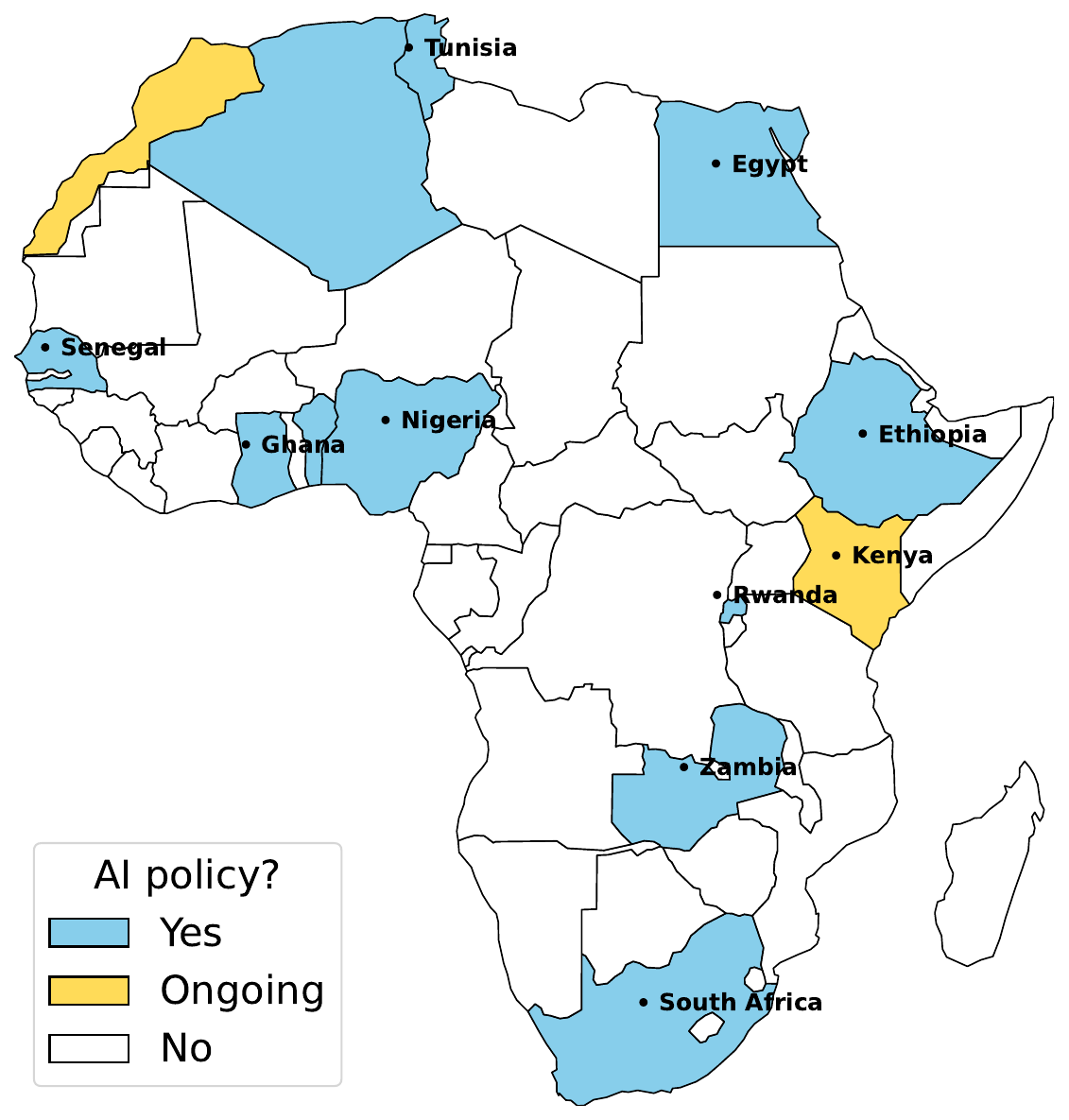}
    \caption{Countries with national AI policy}
    \label{fig:map}
  \end{subfigure}
  \hfill
  \begin{subfigure}[t]{0.59\textwidth}
    \centering
    \includegraphics[height=6.5cm]{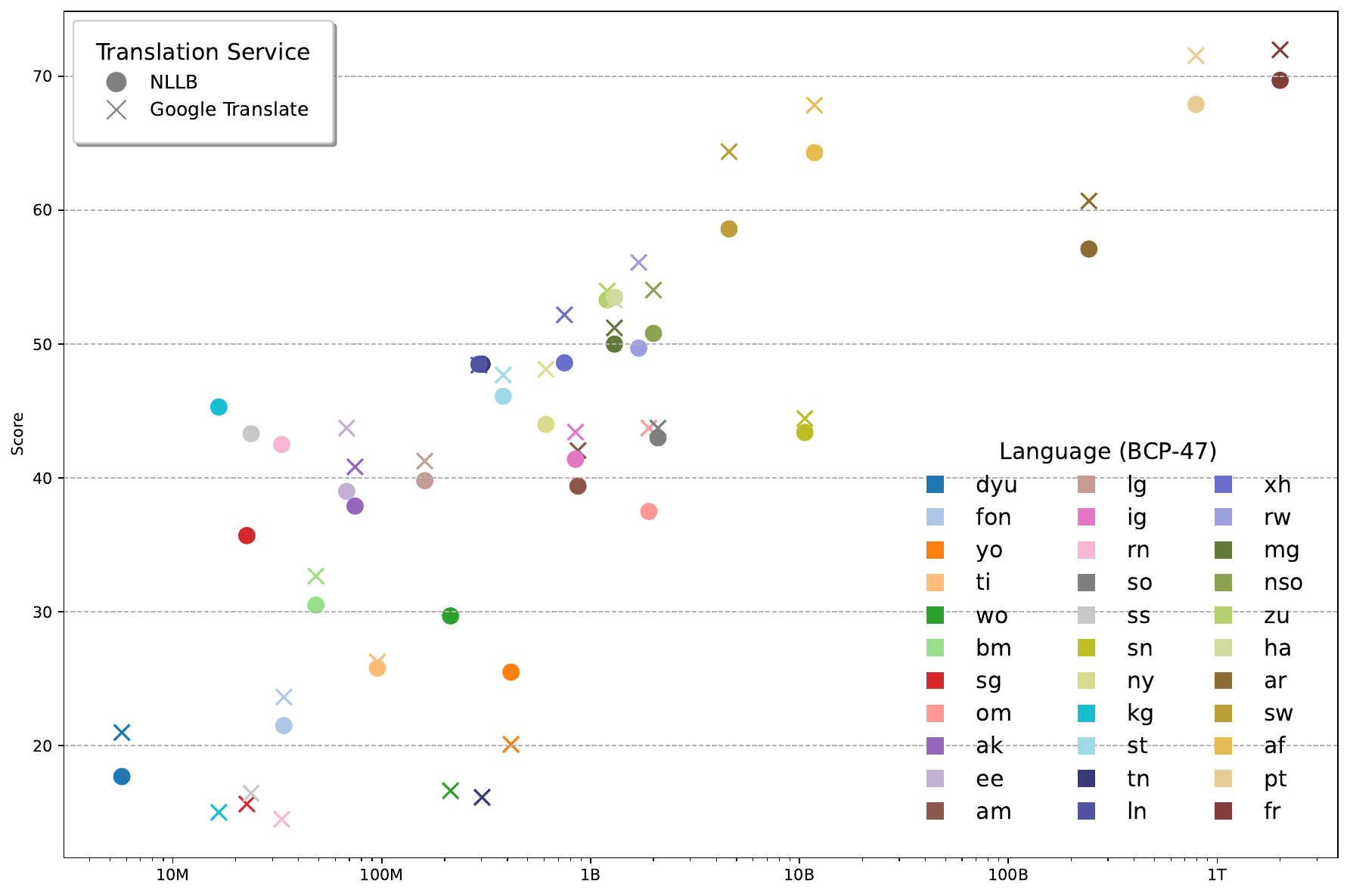}
    \caption{Presence in major AI systems (Chrf2++ vs Characters)}
    \label{fig:image2}
  \end{subfigure}
  \caption{
    \textbf{Overview of African AI Policy and Language Translation Service Presence.} 
    (a) Map showing the status of national AI policies across African countries, categorized as \textit{Yes}, \textit{Ongoing}, and \textit{No}. 
    (b) Scatter plot illustrating the performance of NLLB and Google Translate scores against language character counts (log scale), color-coded by language (BCP-47).
    }
  \label{fig:policy_and_nllbgc}
\end{figure}

\Cref{fig:map} shows a colored map of Africa highlighting countries with national AI policies and those with policies in progress. This reflects a skewed distribution of efforts, indicating that much more remains to be done. However, of the 57 African languages in FLORES, 41 are spoken in at least one country with a national AI policy, a few in countries with policies in progress, and some in countries with no policy at all, such as Kabuverdianu, Ganda, Kikongo, and Tamasheq.

\subsection{Community Input observation }

To better understand the real-world usage and community perspectives on AI readiness across African countries, we designed and distributed a structured Google Forms questionnaire. The form included both multiple-choice and open-ended questions targeting language accessibility, tool availability, and national AI infrastructure. Responses collected from diverse participants across several countries provided valuable insight into the current state of digital inclusion and AI engagement in the region.

From the submissions, we observed that access to digital content in native African languages remains limited. Most respondents found it difficult to locate online resources in their local languages, with only a few describing the experience as “somewhat easy.” Translation and language processing tools were noted to be scarce or underperforming. Google Translate was the most commonly referenced tool, but its effectiveness was often questioned. Community-led efforts such as Nkowaokwu for Igbo, the Academy of Ewe Language and Masakhane Community show promise but are still in early stages and not widely integrated into AI applications.

Additionally, the survey revealed a general lack of awareness or infrastructure surrounding AI development at the national level. Most participants were unsure or unaware of their country’s AI strategies, government investments, or regulatory frameworks. Government support was generally viewed as minimal or unclear. While a few respondents mentioned the existence of university departments or research centers, very few could cite active collaborations between academia and industry. These findings highlight an important gap between grassroots interest and institutional readiness, underscoring the need for more coordinated, inclusive AI development efforts across the continent.

\section{Recommendations and Conclusion}
The objective of this report is to identify African languages that are strong candidates for inclusion in AI safety benchmarks, highlight the types of resources currently missing to support their readiness, and outline the gaps and efforts needed to prepare these languages for meaningful participation in AI safety discussions. In this report, we have assessed the status of these languages based on resource availability, the performance of LLMs, using English as a baseline for comparison, and the presence of national AI policies in the countries where these languages are spoken.

Based on our analyses and findings, we classified languages into four readiness categories: \textbf{fully ready}, \textbf{mostly ready}, \textbf{emerging}, and \textbf{foundational support needed}. Languages in the \textbf{foundational support needed} category include all those not represented in FLORES-200, except the Nigerian Pidgin. Although some of these languages may appear in large multilingual corpora, the lack of standardized evaluation benchmarks to assess their performance relative to other languages supports their placement in this category.

Accordingly, the 58 African languages including the 57 in FLORES-200 were divided into the three remaining categories: we recommend 15 languages as \textbf{fully ready}, 10 languages as \textbf{mostly ready}, while the remaining 33 are categorized as \textbf{emerging}. We compared their presence in Fineweb2, Gemini's performance on translation tasks from English or French to these languages (using FLORES-200), the speaking population, and the existence of a national AI policy in countries where the languages are spoken. \Cref{tab:african_language_categories} presents the final recommendations.

\begin{table}[h!]
\centering
\begin{tabular}{p{4.0cm}|p{11.0cm}}
\toprule
\textbf{Category} & \textbf{Languages} \\
\hline
Fully Ready & Hausa, Egyptian Arabic, Afrikaans, Nigerian Pidgin, Zulu, Swahili, Amharic, Somali, Shona, Nyanja, Southern Sotho, Yoruba, Igbo, Northern Sotho, Kinyarwanda \\
\hline
Mostly Ready & Tsonga, Xhosa, West Central Oromo, Tswana, Rundi, Twi, Bemba, Tunisian Arabic, Tumbuka, Swati \\
\hline
Emerging & Akan, Wolof, Tigrinya, Kabyle, Mossi, Bambara, Moroccan Arabic, Nigerian Fulfulde, Central Kanuri (Latin script), Chokwe, Ewe, Kabiyè, Lingala, Central Kanuri (Arabic script), Luo, Fon, Kikuyu, Central Atlas Tamazight, Nuer, N'ko, Kamba, Kabuverdianu, Plateau Malagasy, Ganda, Luba-Kasai, Sango, Kikongo, Umbundu, Kimbundu, Dyula, Southwestern Dinka, Tamasheq (Latin script), Tamasheq (Tifinagh script) \\
\hline
Foundational support needed &  All African languages not included in FLORES-200, such as [e.g., Tigré, Wolaytta, etc.]  \\
\bottomrule
\end{tabular}
\caption{Categorization of African Languages for AI Safety. the languages are arranged in no particular order.}
\label{tab:african_language_categories}
\end{table}

Aside from the above language categorization and recommendation, we highlight below some activities and effort required for AI safety readiness in African languages.

\begin{enumerate}
    \item \textbf{Strategic Policy Action for Language-Inclusive AI} Governments and policymakers should develop comprehensive national AI policies that explicitly prioritize language inclusion. They should allocate targeted resources, provide grants, and create incentives to support local talent working on language technologies. Their leadership is crucial to ensure that local and underrepresented languages are not left behind in AI advancements and that linguistic diversity is embedded in the nation's digital future.
    
    \item \textbf{Community-Driven Language Inclusion} Local communities and language speakers should take stewardship of collecting clean, verified monolingual data that embodies their cultural values, knowledge systems, and social norms. Their role extends beyond data collection, they are also key advocates for raising awareness about the importance of language representation in AI. This is vital not only for preserving linguistic identity, but also for empowering their communities to participate meaningfully in the AI-driven economy.
    
    \item \textbf{Ensuring AI Safety Through Inclusive Research} Research and Academic Institutions should play a leading role in ensuring AI safety by creating quality evaluation datasets for African languages. These datasets are essential for assessing how large language models (LLMs) perform across different linguistic and cultural contexts. By identifying biases, inaccuracies, and performance gaps, researchers can help prevent harmful outputs and ensure that AI systems behave reliably and ethically for all language communities—not just those with abundant data resources.

\end{enumerate}

 \section*{Acknowledgments}
 This project was supported in MLCommons through a grant with Masakhane Research Foundation.   

\bibliographystyle{unsrt}  
\bibliography{references}

\end{document}